\documentclass[10pt,twocolumn,letterpaper]{article}

\usepackage[]{cvpr}              

\usepackage{lineno}
\usepackage{xcolor}
\usepackage{booktabs}
\usepackage{multirow}    
\usepackage{makecell}
\usepackage{amssymb}
\usepackage{pgfplots}
\pgfplotsset{compat=1.18}
\usepackage{textcomp}
\usepackage{pifont}
\newcommand{\cmark}{\ding{51}} \newcommand{\xmark}{\ding{55}}

\definecolor{cvprblue}{rgb}{0.21,0.49,0.74}
\usepackage[pagebackref,breaklinks,colorlinks,allcolors=cvprblue]{hyperref}

\def\paperID{9055} 
\def\confName{CVPR}
\def\confYear{2026}

\title{GeoGuide: Hierarchical Geometric Guidance for\\ Open-Vocabulary 
3D Semantic Segmentation}

\author{
  Xujing Tao$^{1}$ \hspace{0.5em}
  Chuxin Wang$^{1}$ \hspace{0.5em}
  Yubo Ai$^{1}$ \hspace{0.5em}
  Zhixin Cheng$^{1}$ 
  \hspace{0.5em}
  Zhuoyuan Li$^{1}$ \\
  Liangsheng Liu$^{1}$ \hspace{0.5em}
  Yujia Chen$^{1}$ \hspace{0.5em}
  Xinjun Li$^{1}$ \hspace{0.5em}
  Qiao Li$^{1}$ \hspace{0.5em}
  Wenfei Yang$^{1}$ \hspace{0.5em}
  Tianzhu Zhang$^{1,2}$\thanks{Corresponding author.}
  \vspace{0.5em} \\
  $^1$University of Science and Technology of China \\
  $^2$National Key Laboratory of Deep Space Exploration, Deep Space Exploration Laboratory \\
}

\begin{document}
\maketitle
\begin{abstract}
Open-vocabulary 3D semantic segmentation aims to segment arbitrary categories beyond the training set.
Existing methods predominantly rely on distilling knowledge from 2D open-vocabulary models. However, aligning 3D features to the 2D representation space restricts intrinsic 3D geometric learning and inherits errors from 2D predictions. To address these limitations, we propose \textbf{GeoGuide}, a novel framework that leverages pretrained 3D models to integrate hierarchical geometry-semantic consistency for open-vocabulary 3D segmentation. 
Specifically, 
we introduce an \textbf{Uncertainty-based Superpoint Distillation} module to fuse geometric and semantic features for estimating per-point uncertainty, adaptively weighting 2D features within superpoints to suppress noise while preserving discriminative information  to enhance local semantic consistency. Furthermore, our \textbf{Instance-level Mask Reconstruction} module leverages geometric priors to enforce semantic consistency within instances by reconstructing complete instance masks. 
Additionally, our \textbf{Inter-Instance Relation Consistency} module aligns geometric and semantic similarity matrices to calibrate cross-instance consistency for same-category objects, mitigating viewpoint-induced semantic drift.
Extensive experiments on ScanNet v2, Matterport3D, and nuScenes demonstrate the superior performance of GeoGuide.
\end{abstract}

\section{Introduction}
\label{sec:intro}

3D scene understanding is critical for applications such as autonomous driving~\cite{ettinger2021large,caesar2020nuscenes} and embodied intelligence~\cite{seita2023toolflownet}.
As a core subtask, point cloud semantic segmentation aims to predict a category label for each point in 3D space. 
While fully-supervised methods have achieved great success, they rely on densely annotated 3D datasets and inherently fail to handle unseen categories in real-world scenarios.

\begin{figure}
    \centering
    \includegraphics[width=1.0\linewidth]{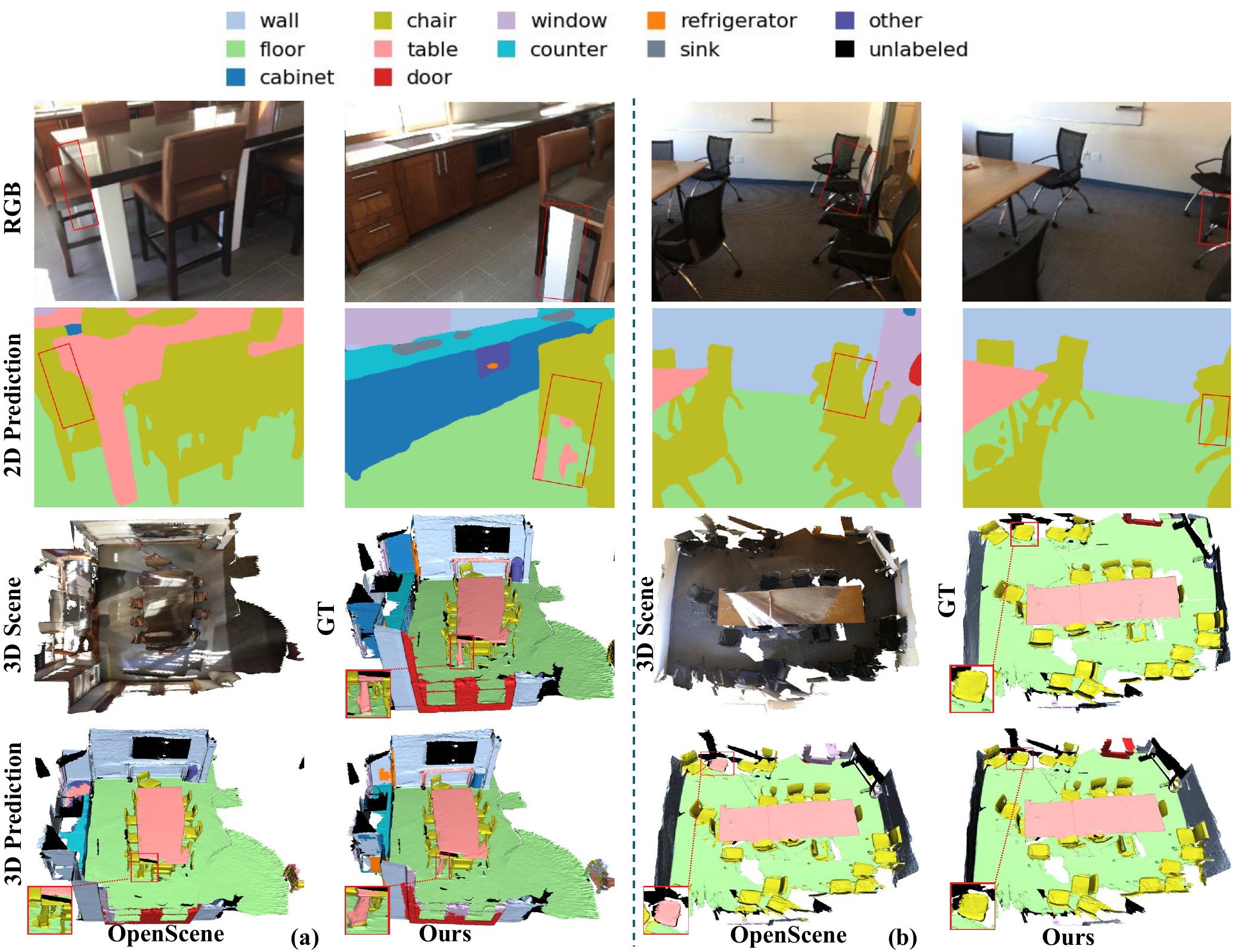}
    \vspace{-1.9em}
    \caption{
    \textbf{(a)} Occlusions and ambiguous boundaries in 2D predictions lead to inaccurate 3D segmentation, whereas our geometry-guided modeling effectively rectifies these errors.
    \textbf{(b)} The baseline suffers from the lack of geometric relation modeling, whereas our method exploits inter-instance geometric constraints to produce accurate predictions.}
    \vspace{-1.7em}
    \label{fig:one}
\end{figure}

To address this limitation, 3D open-vocabulary semantic segmentation has attracted significant attention, enabling models to segment 3D scenes based on arbitrary natural language descriptions. 
However, unlike 2D open-vocabulary tasks that benefit from abundant large-scale image-text pairs~\cite{radford2021learning,cheng2026adaptive}, 3D point-text paired data remain scarce due to prohibitive annotation costs~\cite{ding2023pla,lee2025mosaic3d,cheng2025bridge,cheng2025i2p,cheng2025b2}.
Consequently, recent works tackle this challenge by transferring knowledge from pre-trained 2D open-vocabulary models.
These methods generally follow two paradigms.
(1) 2D-to-3D distillation~\cite{peng2023openscene,li2025sas,wang2024open,li2024dense,zhang2023clip,cheng4,cheng6,cheng7,cheng8,cheng9}, which projects pixel-level semantic features onto point clouds through geometric correspondences and aligns 3D representations with 2D features~\cite{li2022language,ghiasi2022scaling} via knowledge distillation.
(2) Point–text alignment~\cite{jiang2024open,ding2023pla,lee2025mosaic3d,yang2024regionplc,1,2,3,4,5,6}, which generates textual descriptions from images, constructs point–text pairs using geometric correspondences, and aligns 3D features with textual embeddings through contrastive learning.
Essentially, both paradigms train the 3D model to replicate the feature representations of 2D models, which restricts the learning of intrinsic 3D geometric structures.
It is worth noting that 2D models are prone to producing erroneous object masks due to occlusions and viewpoint changes, causing the 3D model to inherit these errors and learn incorrect segmentation patterns, as shown in Figure~\ref{fig:one}.

Based on the above analysis, we argue that the core limitation of existing open-vocabulary 3D segmentation methods lies in \textbf{how to effectively preserve intrinsic 3D geometric information} during the 2D-to-3D knowledge distillation process.
A natural idea is to leverage pretrained 3D models to correct the geometric biases from 2D representations.
However, directly incorporating pretrained 3D features does not necessarily improve performance, as the heterogeneous supervision signals between different modalities introduce instability during training.
To address this challenge, we identify three key issues that must be resolved to achieve stable and geometry-consistent open-vocabulary 3D segmentation.
%
%
\textbf{(1) Intra-Superpoint Consistency Modeling.}
Superpoints~\cite{papon2013voxel}, defined as geometrically homogeneous neighborhoods, should inherently share the same semantic label.
However, when projecting 2D predictions onto 3D, semantics within a single superpoint often become inconsistent, particularly due to 2D-level occlusions or boundary ambiguities.
Existing approaches typically enforce consistency through mean pooling, which suppresses discriminative features and is overly sensitive to prediction noise, thereby amplifying deviations from the correct semantic representation.
Hence, maintaining reliable intra-superpoint semantic consistency remains a fundamental challenge.
\textbf{(2) Intra-Instance Consistency Modeling.} 
Predictions from a single 2D view are inherently incomplete, capturing only partial masks that fail to represent the full 3D geometry of each instance.
This incompleteness leads to fragmented instance-level semantics in 3D space.
Thus, achieving consistent semantics across all parts of an instance and reconstructing its complete structure remains a key challenge in 2D-to-3D distillation.
\textbf{(3) Inter-Instance Relation Consistency Modeling.} 
Aggregating 2D features from multiple views introduces feature distribution shifts among same-category instances due to viewpoint variations.
As a result, instances of the same class may exhibit divergent feature representations, leading to semantic inconsistency across the scene.
Therefore, modeling geometric relation constraints to ensure consistent semantics among same-category instances poses another open challenge.

To tackle these challenges, we propose a hierarchical geometry-guided framework that explicitly integrates geometric priors at multiple levels to achieve fine-grained geometry–semantic alignment.
The framework consists of three key modules: the Uncertainty-based Superpoint Distillation (USD) module, the Instance-level Mask Reconstruction (IMR) module, and the Inter-Instance Relation Consistency (IIRC) module.
%
%
\textbf{In the USD module}, we fuse geometric features from a pretrained 3D backbone with 2D semantic features to estimate per-point feature uncertainty.
This uncertainty guides weighted aggregation within each superpoint to highlight discriminative features while suppressing noise. 
Through superpoint-level 2D-to-3D feature distillation, this module enforces robust intra-superpoint semantic consistency.
\textbf{In the IMR module}, we generate class-agnostic instance proposals following~\cite{yin2024sai3d} and reconstruct complete instance masks to recover missing regions, achieving intra-instance consistency by unifying the semantics of all points within an instance.
\textbf{In the IIRC module}, we exploit the observation that pretrained 3D models produce similar geometric representations for same-category objects.
By aligning geometric and semantic similarity matrices across instances, we ensure inter-instance consistency, mitigating semantic drift caused by viewpoint variation.
This relation consistency is jointly enforced at both the superpoint and instance levels, forming a local-to-global semantic alignment mechanism.
%

%
Our main contributions are summarized as follows:
(1) We identify the limitations of conventional 2D feature distillation in open-vocabulary 3D segmentation and present a novel framework that leverages pretrained 3D models to integrate hierarchical geometric priors for enhanced geometric awareness.
(2) We design three complementary modules that enforce geometry–semantic consistency at the intra-superpoint, intra-instance, and inter-instance levels, achieving robust local-to-global semantic alignment.
(3) Extensive experiments on multiple benchmarks demonstrate that our method achieves state-of-the-art performance and exhibits strong cross-domain generalization, while comprehensive ablation studies validate the effectiveness of each proposed component.

\section{Related Work}
\subsection{Closed-Set 3D Scene Understanding}
3D scene understanding underpins applications in autonomous driving, embodied intelligence, and robotic navigation. 
Significant progress has been made in various 3D vision tasks, including classification~\cite{chen2023clip2scene,qi2017pointnet,li2025pamba,deng2023se}, semantic segmentation~\cite{qi2017pointnet++,wu2024point,deng2025quantity,landrieu2018large,zhao2021point}, object detection~\cite{lang2019pointpillars,pan20213d,zhou2018voxelnet,deng2024diff3detr,wang2025state,wang2023long,lu2025trackingworld,wang2023not,lu2024bsnet}, and instance segmentation~\cite{jiang2020pointgroup,kolodiazhnyi2024oneformer3d,vu2022softgroup,lu2023query,lu2025relation3d,sun2023superpoint}.
However, most methods rely on training with densely annotated datasets, thus being restricted to predefined categories and failing to generalize to novel categories.
In contrast, our work aims to achieve annotation-free 3D open-vocabulary understanding, improving generalization under open-set conditions.

\subsection{Large-Scale Pretrained Models}
Large-scale pretrained models have achieved remarkable success in NLP and 2D vision tasks, such as DINOv3~\cite{simeoni2025dinov3} and MAE~\cite{he2022masked}. Inspired by these advances, numerous 3D pretraining strategies have emerged, typically categorized into contrastive learning and generative modeling.
%
Contrastive learning approaches learn discriminative representations by enforcing consistency across different views or modalities. PointContrast~\cite{xie2020pointcontrast} first extends 2D contrastive learning to 3D domain, while CrossPoint~\cite{afham2022crosspoint} introduces intra-modal and inter-modal contrastive objectives for improved global discrimination.
Generative approaches~\cite{zhu2024motiongs} adopt masked point modeling (MPM) to reconstruct masked point clouds and learn geometric priors. Point-BERT~\cite{yu2022point} encodes local geometric information via discrete tokens, and PointMAE~\cite{pang2023masked} reconstructs masked regions from high-level latent features. Sonata~\cite{wu2025sonata} further enhances representation learning through intra-modal self-distillation.
These pretraining methods capture strong geometric priors from large-scale point cloud data. In this work, we explore how to effectively transfer such pretrained 3D geometric priors to open-vocabulary 3D understanding tasks.

\subsection{Open-Vocabulary 3D Semantic Segmentation}
Open-vocabulary 3D semantic segmentation enables recognition of unseen categories in 3D scenes.
With the emergence of vision-language representation models (VLMs) such as CLIP~\cite{radford2021learning}, substantial progress has been made in 2D open-vocabulary segmentation and detection~\cite{li2022language,ghiasi2022scaling,liang2023open,xu2023side,zou2023segment}.
However, extending open-vocabulary capability to 3D domain remains challenging due to the scarcity of large-scale paired point–text data.
%
To address this, recent works transfer 2D open-vocabulary knowledge to the 3D domain. OpenScene~\cite{peng2023openscene} aligns point clouds with multi-view 2D features using geometric correspondences and distills language-aligned representations into 3D encoders.
Subsequent approaches generate multi-view textual descriptions to align 3D features with text embeddings~\cite{ding2023pla,yang2024regionplc,jiang2024open,wang2025masked,lee2025mosaic3d,zhang2025pgov3d,mei2024geometrically,lu2024dn,yang2023sam3d,wang2024uniplv,wang2024xmask3d,he2024unim}.
GGSD~\cite{wang2024open} adopts a mean-teacher framework to further enhance feature distillation. SAS~\cite{li2025sas} and CUA-O3D~\cite{li2025cross} integrate multiple 2D open-vocabulary models to reduce individual model bias.
Despite these advances, most approaches remain overly reliant on 2D supervision and overlook the intrinsic 3D geometric structure, leading to inconsistent or geometry-agnostic representations.
In contrast, our method explicitly incorporates a pretrained 3D model and leverages its latent geometric priors to enhance geometry–semantic consistency, thereby achieving more accurate and robust open-vocabulary 3D scene understanding.

\begin{figure*}[t]
    \centering
    \includegraphics[width=1.0\linewidth]{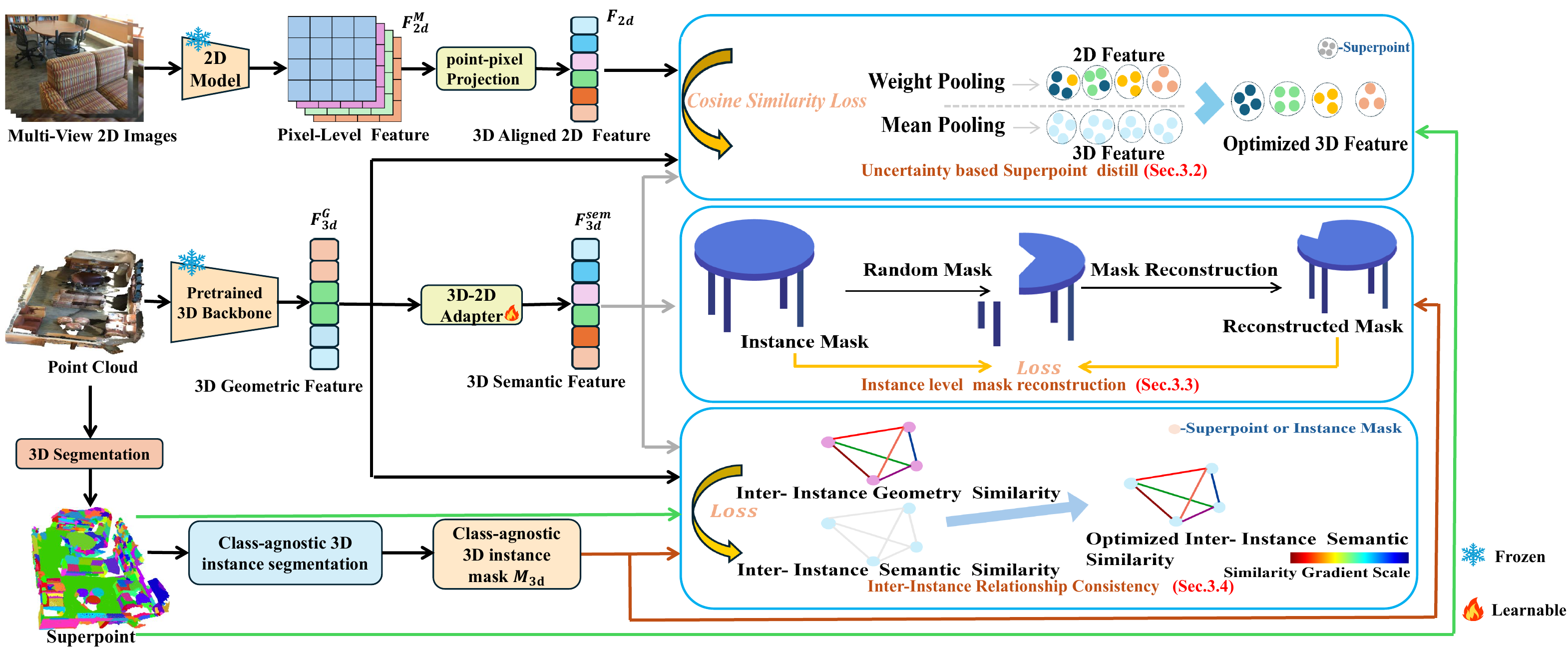}
    \vspace{-2em}
    \caption{\textbf{Overview of our proposed GeoGuide framework}. GeoGuide integrates hierarchical geometric priors to achieve geometry–semantic consistent open-vocabulary 3D segmentation}
   \vspace{-1.0em}
    \label{fig:two}
\end{figure*}


\section{Method}
\subsection{Framework Overview}
\label{sec:framework}
We propose $\textbf{GeoGuide}$, a geometry-guided framework for robust and accurate  open-vocabulary 3D semantic segmentation. The overall architecture is illustrated in Fig~\ref{fig:two}. Given a scene point cloud $\mathbf{P} \in \mathbb{R}^{N \times 3}$ (where $N$ is the number of points) and its corresponding multi-view RGB images $\mathcal{I} \in \mathbb{R}^{M \times 3 \times H \times W}$, we first extract 3D geometric features $\mathbf{F}_{3d}^\text{G} \in \mathbb{R}^{N \times C_1}$ using a pretrained 3D backbone. In parallel, we obtain pixel-level semantic features $\mathbf{F}_{2d}^M \in \mathbb{R}^{M \times H\times W\times C}$ from a frozen 2D open-vocabulary segmentation model.
Using camera parameters and depth maps, we establish geometric correspondences between 3D points and image pixels, projecting 2D  features onto the point cloud to obtain aligned 2D semantic features $\mathbf{F}_{2d} \in \mathbb{R}^{N \times C}$.
The 3D geometric features are then mapped into the same semantic space through a lightweight 3D-2D adapter, producing 3D semantic representations $\mathbf{F}_{3d}^{\text{sem}} \in \mathbb{R}^{N \times C}$.
To effectively integrate geometric priors with semantic knowledge during 2D-to-3D distillation, we introduce three hierarchical geometry-guided modules: Uncertainty-based Superpoint Distillation (Sec.~\ref{sec:usd}), Instance-level Mask Reconstruction (Sec.~\ref{sec:imr}), and Inter-Instance  Relation Consistency (Sec.~\ref{sec:iirc}).
These modules progressively enforce geometry–semantic consistency from local to global levels, guiding the distillation process to preserve geometric structure while learning semantically enriched representations.

\subsection{Feature Extraction and \text{2D-3D} Alignment}
\label{sec:feature_extraction}
\noindent\textbf{Geometric and Semantic Feature Extraction.}
We employ a frozen 3D backbone~\cite{wu2025sonata} pretrained on large-scale 3D datasets to extract geometric features  $\mathbf{F}_{3d}^\text{G} \in \mathbb{R}^{N \times C_1}$,  In parallel, frozen 2D open-vocabulary segmentation models generate vision–language aligned pixel-level features from 
 multi-view RGB images 
 , denoted as  $\mathbf{F}_{2d}^M \in \mathbb{R}^{M \times H\times W\times C}$. 

\noindent\textbf{2D-3D Correspondence Establishment.}
Each 3D point $\mathbf{P}_i \in \mathbf{P}$ is projected onto multi-view images using intrinsic and extrinsic matrices to establish 2D–3D correspondences: 
\begin{equation}
\vspace{-0.4em}
\label{eq:projection}
d_{i,t} \cdot \begin{bmatrix} u_i \\ v_i \\ 1 \end{bmatrix}_t = \mathbf{\Gamma} \cdot [\mathbf{R}|\mathbf{t}]_t \cdot \begin{bmatrix} x_i \\ y_i \\ z_i \\ 1 \end{bmatrix},
\vspace{-0.5em}
\end{equation}

Projection pairs are retained only if: (1) the pixel  coordinates $(u_i,v_i)$ lie  within image boundaries and (2) the depth difference between  $d_{i,t}$ (projected depth) and $D_t$ (sensor-captured depth) is below a threshold, i.e.,$|d_{i,t} - D_t(\lfloor u_i \rfloor, \lfloor v_i \rfloor)| < \tau_{\text{depth}}$.  Each point may correspond to multiple valid projections across views, following  \cite{peng2023openscene}, the projected features are fused via average pooling $\mathbf{f}_{2d} = \phi(\mathbf{f}_1, \cdots, \mathbf{f}_K)$ to form $\mathbf{F}_{2d} \in \mathbb{R}^{N \times C}$.
%

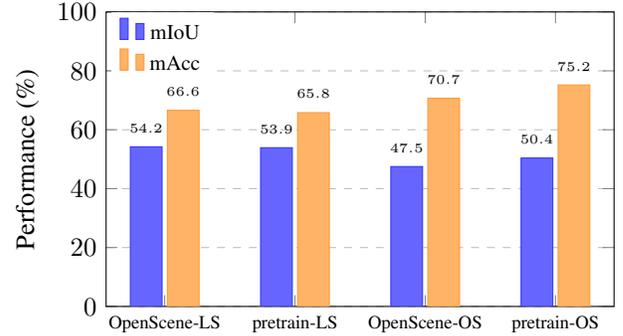
\begin{figure}[t]
    \centering
    \begin{tikzpicture}
        \begin{axis}[
            width=\columnwidth,
            height=5.5cm,
            ybar,
            bar width=12pt,
            ylabel={Performance (\%)},
            symbolic x coords={OpenScene-LS, pretrain-LS, OpenScene-OS, pretrain-OS}, 
            xtick=data,
            x tick label style={font=\scriptsize},
            ymin=0, ymax=100,
            ytick={0,20,40,60,80,100},
            ymajorgrids=true,
            grid style=dashed,
            legend style={at={(0.01,1)}, anchor=north west, font=\footnotesize,draw=none},
            enlarge x limits=0.15,
            nodes near coords,
            nodes near coords style={font=\tiny, yshift=2pt},
             tick align=inside,
        ]
        
        \addplot[fill=blue!60, draw=blue!80] coordinates {
            (OpenScene-LS, 54.2)
            (pretrain-LS, 53.9)  
            (OpenScene-OS, 47.5)  
            (pretrain-OS, 50.4)
        };
        
        \addplot[fill=orange!60, draw=orange!80] coordinates {
            (OpenScene-LS, 66.6)
            (pretrain-LS, 65.8)  
            (OpenScene-OS, 70.7)  
            (pretrain-OS, 75.2)
        };
        
        \legend{mIoU, mAcc}
        \end{axis}
    \end{tikzpicture}
    \vspace{-1.0em}
    \caption{\textbf{Preliminary Experiments.} Comparison between OpenScene and our preliminary attempt under different 2D models. LS: LSeg, OS: OpenSeg.}
    \vspace{-1.2em}
    \label{fig:preliminary_exp}
\end{figure}

\noindent\textbf{Preliminary Experiments.} We project the geometric features $\mathbf{F}_{3d}^{\text{G}} \in \mathbb{R}^{N \times C_1}$ into the vision-language space using a  two-layer MLP adapter, obtaining  $\mathbf{F}_{3d}^{\text{sem}} = {VL}(\mathbf{F}_{3d}^{\text{G}})$, where ${VL}: \mathbb{R}^{N \times C_1} \to \mathbb{R}^{N \times C}$.
%
As shown in Fig.~\ref{fig:preliminary_exp}, we compare OpenScene's direct 2D-to-3D distillation (which trains the entire 3D network) with our preliminary attempt, which freezes the pretrained 3D backbone~\cite{wu2025sonata} and only trains a lightweight MLP adapter. Both methods distill knowledge from 2D open-vocabulary models (LSeg\cite{li2022language} and OpenSeg\cite{ghiasi2022scaling}) into 3D models. Results reveal inconsistent performance gains across different 2D models: with OpenSeg, freezing the pretrained backbone significantly improves performance (mIoU: 47.5$\to$50.4, mAcc: 70.7$\to$75.2), whereas with LSeg, the improvement is marginal or slightly negative (mIoU: 54.2$\to$53.9). This inconsistency indicates that naive 2D-to-3D distillation can disrupt geometric priors learned during 3D pretraining. We identify two key reasons: First, 3D geometric features and 2D texture-based features belong to fundamentally different modalities with potentially significant feature-space discrepancies, leading to inconsistent learning signals during alignment. Second, 2D models are susceptible to occlusions, illumination variations, and viewpoint variations, leading to erroneous predictions.
To effectively integrate 3D geometric priors and 2D semantic information, we design three hierarchical geometric-semantic consistency modules that range from local to global to guide the  distillation process, ensuring robust and geometry-aware 3D open-vocabulary segmentation.

\subsection{Hierarchical Geometry-Guided Modules}
\label{sec:modules}
\subsubsection{Uncertainty-Based Superpoint Distillation}
\label{sec:usd}
\begin{figure}
    \centering
    \includegraphics[width=1.0\linewidth]{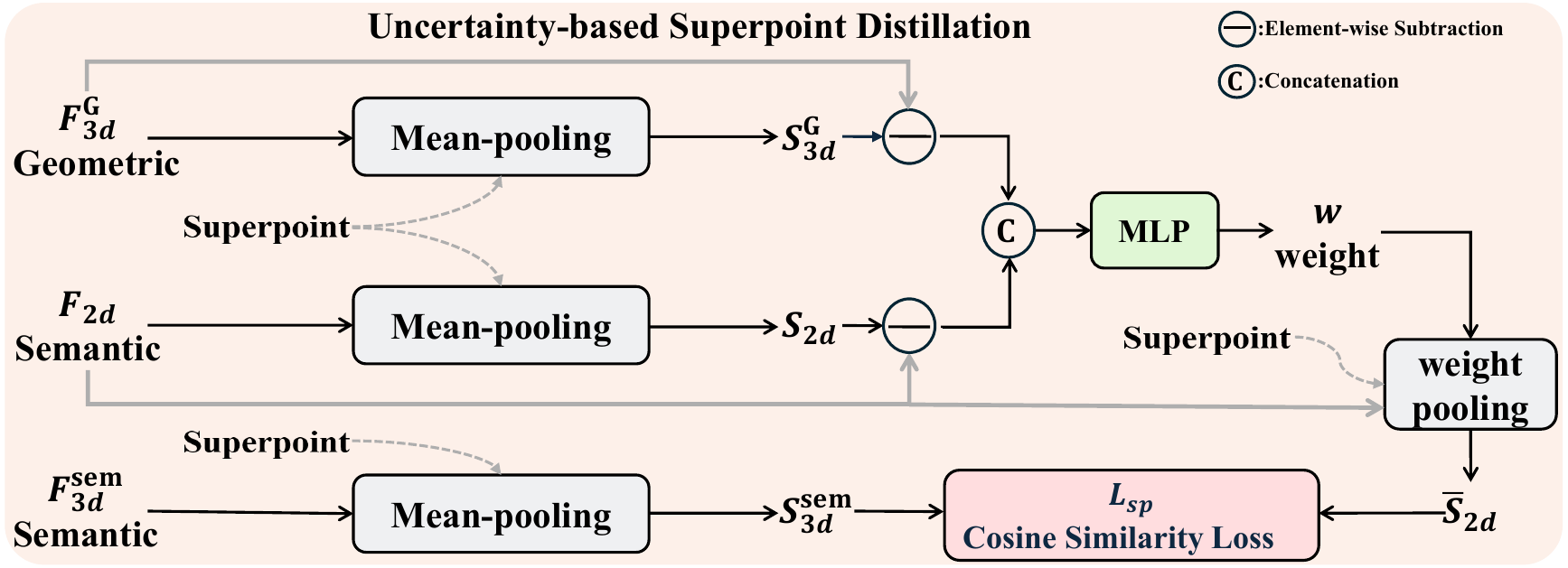}
    \vspace{-1.5em}
    \caption{\textbf{Overview of the USD}. Uncertainty prediction from 3D geometry and 2D semantics guides the  distillation of semantic knowledge.}
    \vspace{-1.0em}
    \label{fig:three}
\end{figure}

This module utilizes geometric consistency within superpoints to promote semantic coherence, as points grouped into the same superpoint are expected to belong to the same semantic category.
By combining geometric and semantic information, we estimate uncertainty weights for the 2D features corresponding to each point, enabling adaptive weighted aggregation of superpoint features that emphasizes discriminative and correct 2D semantic features while suppressing irrelevant noisy features. As shown in Fig.~\ref{fig:three}, we first apply normal-based over-segmentation to the point cloud, yielding superpoints $\{Q_i\}_{i=1}^{N_Q}$. We then perform mean pooling of $\mathbf{F}_{3d}^{\text{sem}}$, $\mathbf{F}_{3d}^\text{G}$, and $\mathbf{F}_{2d}$ within each superpoint to obtain their respective superpoint-level features $\mathbf{S}_{3d}^{\text{sem}} \in \mathbb{R}^{N_Q \times C}$, $\mathbf{S}_{3d}^{\text{G}} \in \mathbb{R}^{N_Q \times C_1}$ and $\mathbf{S}_{2d} \in \mathbb{R}^{N_Q \times C}$. We compute the differences between the superpoint-level features and their corresponding point-level features. These different features are then concatenated and fed through a single-layer MLP to predict reliability weights $\mathcal{W}$ for the 2D semantic features of each point. 
\begin{equation}
\label{eq:uncertainty}
\mathcal{W} = \text{MLP}(\text{concat}[(\mathbf{S}_{3d}^{\text{G}} - \mathbf{F}_{3d}^{\text{G}}); (\mathbf{S}_{2d} - \mathbf{F}_{2d})]) \in \mathbb{R}^{N \times 1},
\end{equation}
where $\text{concat}[\cdot;\cdot]$ denotes the concatenation operation. 
Using these predicted weights, we perform weighted pooling on $\mathbf{F}_{2d}$ to obtain refined superpoint-level 2D semantic features $\overline{\mathbf{S}}_{2d} \in \mathbb{R}^{N_Q \times C}$, which are then mapped back to point-level features $\overline{\mathbf{F}}_{2d} \in \mathbb{R}^{N \times C}$:
\begin{equation}
\label{eq:weighted_pooling}
\overline{\mathbf{S}}_{2d} = \text{WP}(\mathcal{W}, \mathbf{F}_{2d},\{Q_i\}_{i=1}^{N_Q} ),
\end{equation}
where $\text{WP}(\cdot,\cdot)$ denotes weighted pooling. We then compute cosine-similarity-based distillation losses at both superpoint and point levels:
\begin{equation}
\label{eq:loss_sp}
\mathcal{L}_{\text{sp}} = \left(1 - \cos(\mathbf{S}_{3d}^{\text{sem}}, \overline{\mathbf{S}}_{2d})\right) + \left(1 - \cos(\mathbf{F}_{3d}^{\text{sem}}, \overline{\mathbf{F}}_{2d})\right).
\end{equation}

\subsubsection{Instance-Level Mask Reconstruction}
\label{sec:imr}
The superpoints generated in Sec.~\ref{sec:usd} typically cover only partial regions of instances, so the USD module enforces semantic consistency mainly within local instance regions. 
To enforce semantic consistency at the instance level, we leverage a class-agnostic 3D instance segmentation method~\cite{yin2024sai3d} to obtain a set of instance masks $\{M_i\}_{i=1}^{N_M}$. The IMR module then operates on these masks to enforce global instance-level consistency.  For each mask $M_i$, we randomly mask out a portion to obtain an incomplete mask $\overline{M}_i$. We then use $\overline{M}_i$ to index the corresponding point features in $\mathbf{F}_{3d}^{\text{sem}} $, apply average pooling, and pass the result through a linear layer to obtain the mask feature $\overline{\mathbf{F}}_i^{\text{mask}} \in \mathbb{R}^{1 \times C}$:
\begin{equation}
\label{eq:mask_feature}
\overline{\mathbf{F}}_i^{\text{mask}} = \text{Linear}(\text{pool}(\mathbf{F}_{3d}^{\text{sem}}[\overline{M}_i])),
\end{equation}
where $\text{Linear}: \mathbb{R}^{C} \to \mathbb{R}^C$. We then predict the reconstructed mask $\hat{M}_i$ using $\overline{\mathbf{F}}_i^{\text{mask}}$ and $\mathbf{F}_{3d}^{\text{sem}}$:

\begin{equation}
\label{eq:mask_reconstruction}
\hat{M}_i = \text{sigmoid}(\cos(\overline{\mathbf{F}}_i^{\text{mask}}, \mathbf{F}_{3d}^{\text{sem}})).
\end{equation}
We employ a binary cross-entropy (BCE) loss to measure the difference between the reconstructed mask and the original mask:

\begin{equation}
\label{eq:loss_mask}
\mathcal{L}_{\text{mask}} = \frac{1}{N_M}\sum_{i=1}^{N_M}\text{BCE}(M_i, \hat{M}_i).
\end{equation}
This loss encourages points within the same instance to learn similar semantic features.

\subsubsection{Inter-Instance Relation Consistency}
\label{sec:iirc}
Aggregating 2D features from multiple views often introduces feature distribution shifts among instances of the same category due to viewpoint variations and occlusions, leading to semantic divergence. Pretrained 3D backbones capture strong geometric priors that ensure same-category objects have similar geometric representations. However, these priors are not automatically preserved during 2D-to-3D distillation. 
To address this, we propose the Inter-Instance Relation Consistency (IIRC) module to align semantic relationships with geometric affinities across instances. The key insight is that the geometric similarity matrix, which encodes pairwise structural relationships among instances in 3D space, should be reflected in the semantic similarity matrix of the distilled 3D semantic features. By enforcing the semantic similarity to match the geometric similarity, our module guides the semantic features to respect the intrinsic geometric structure and mitigates semantic drift caused by viewpoint variations. Specifically, given instance masks $\{M_i\}_{i=1}^{N_M}$,
we aggregate the corresponding 3D geometric features $\mathbf{F}_{3d}^{\text{G}}$ and semantic features $\mathbf{F}_{3d}^{\text{sem}}$ into mask-level embeddings: $\mathbf{F}_{\text{mask}}^{\text{G}} \in \mathbb{R}^{N_M \times C}$ and $\mathbf{F}_{\text{mask}}^{\text{sem}} \in \mathbb{R}^{N_M \times C}$. Pairwise similarity matrices are then computed as:
\begin{equation}
\label{eq:mask_similarity}
\mathbf{P}_{\text{sim-m}}^{\text{G}} = \mathbf{F}_{\text{mask}}^{\text{G}} {\mathbf{F}_{\text{mask}}^{\text{G}}}^{\mathrm{T}}, \quad \mathbf{P}_{\text{sim-m}}^{\text{sem}} = \mathbf{F}_{\text{mask}}^{\text{sem}} {\mathbf{F}_{\text{mask}}^{\text{sem}}}^{\mathrm{T}}.
\end{equation}
\noindent Similarly, we compute the inter-superpoint geometric similarity matrix
$\mathbf{P}_{\text{sim-sp}}^{\text{G}} \in \mathbb{R}^{N_Q \times N_Q}$ and semantic similarity matrix $\mathbf{P}_{\text{sim-sp}}^{\text{sem}} \in \mathbb{R}^{N_Q \times N_Q}$ using superpoint-level geometric features $\mathbf{S}_{3d}^{\text{G}}$ and semantic features $\mathbf{S}_{3d}^{\text{sem}}$. We employ the mean squared error (MSE) loss to align the semantic similarity matrices with their geometric counterparts. The overall loss $\mathcal{L}_{sim}$ for this module is defined as: 

\begin{equation}
\label{eq:loss_sim}
\mathcal{L}_{\text{sim}} = \text{MSE}(\mathbf{P}_{\text{sim-m}}^{\text{G}}, \mathbf{P}_{\text{sim-m}}^{\text{sem}}) + \text{MSE}(\mathbf{P}_{\text{sim-sp}}^{\text{G}}, \mathbf{P}_{\text{sim-sp}}^{\text{sem}}),
\end{equation}
which constrains the semantic similarity matrices to align with the geometric similarity matrices, preventing the degradation of geometric consistency information learned during pretraining across different instances of the same category. This ensures that the distilled point cloud semantic features $\mathbf{F}_{3d}^{\text{sem}}$ maintain semantic consistency across different instances of the same category.



\subsection{Model Training and Inference}
\label{sec:training}
The final loss function during training is defined as:

\begin{equation}
\label{eq:loss_final}
\mathcal{L}_{\text{final}} = \lambda_1\mathcal{L}_{\text{sp}} + \lambda_2\mathcal{L}_{\text{mask}} + \lambda_3\mathcal{L}_{\text{sim}},
\end{equation}

\noindent where $\lambda_1$ , $\lambda_2$, $\lambda_3$ are hyperparameters.
During inference, we only require the 3D point cloud as input, and we drop out the proposed  three Geometry-Guided Modules. We use the output of the 3D-2D adapter $\mathbf{F}_{3d}^{\text{sem}}$ as the point-level features for inference. Using the CLIP text encoder, we compute text embeddings $\mathbf{T} \in \mathbb{R}^{N_C \times C}$ for any given set of category texts $\{\text{text}_i\}_{i=1}^{N_C}$. Each point is then assigned to the category with the highest similarity by computing the similarity between the point feature $\mathbf{f}_{3d}^\text{sem}$ and the text embeddings $\mathbf{T}$.


\section{Experiments}
\subsection{Experimental Setup}
\textbf{Datasets and Metrics.} 
We conduct comprehensive experiments and ablation studies on three widely used benchmark datasets: ScanNet v2~\cite{dai2017scannet}, Matterport3D~\cite{chang2017matterport3d}, and nuScenes~\cite{caesar2020nuscenes}. 
\textbf{ScanNet v2} is an indoor RGB-D dataset with 1,613 scenes, of which 1,201 for training, 301 for validation, and 100 for testing, annotated with 20 semantic categories. 
\textbf{Matterport3D} comprises 10,800 large-scale indoor building scenes with RGB-D images and 3D meshes annotated with 21 base categories. 
\textbf{nuScenes} is an outdoor autonomous driving dataset consisting of 1,000 LiDAR sequences covering 23 semantic categories. 
We follow the official dataset splits and report results on the validation set for ScanNet v2 and nuScenes, and on the test set for Matterport3D. The evaluation metrics include mean Intersection over Union (mIoU) and mean Accuracy (mAcc).

\noindent\textbf{Implementation Details.}
Our framework is implemented in PyTorch and all experiments are conducted on a single NVIDIA A6000 GPU.
We adopt the pretrained 3D backbone from~\cite{wu2025sonata}. Following  OpenScene~\cite{peng2023openscene}, the voxel size is set to 2\,cm and 5\,cm for indoor and outdoor datasets, respectively. We use AdamW~\cite{loshchilov2017decoupled} as the optimizer with a batch size of 4 for all datasets. For indoor datasets, we adopt  LSeg~\cite{li2022language}, OpenSeg~\cite{ghiasi2022scaling}, and the LSeg+SEEM combination from SAS~\cite{li2025sas}. For outdoor datasets, we use LSeg, OpenSeg, and OpenSeg+SEEM. All reported results of our method are obtained from the \textbf{pure 3D model} and we do not use the time-consuming self-distillation strategies proposed by \cite{wang2024open,li2025sas}. More implementation details are provided in the supplementary material.

\begin{table}
\centering
\small
\setlength{\tabcolsep}{1.8pt}
\renewcommand{\arraystretch}{0.95}
\caption{\textbf{Open-vocabulary 3D semantic segmentation results.} Comparison between \textbf{GeoGuide} and existing fully-supervised and zero-shot methods. $\text{SAS}^*$ denotes using the same 2D features as in SAS~\cite{li2025sas}, $\ddag$ denotes the use of the 2D-3D ensemble strategy, \textbf{LS: LSeg, OS: OpenSeg}. Best results are shown in \textbf{bold}.}
\vspace{-1em}
\label{tab:multi_dataset_compare}
\begin{tabular}{lcccccc}
\toprule
\multirow{2}{*}{{Method}} & \multicolumn{2}{c}{ScanNetv2} & \multicolumn{2}{c}{nuScenes} & \multicolumn{2}{c}{Matterport3D} \\
\cmidrule(lr){2-3} \cmidrule(lr){4-5} \cmidrule(lr){6-7} 
& mIoU & mAcc & mIoU & mAcc & mIoU & mAcc \\
\midrule
\multicolumn{7}{c}{\textit{Fully-supervised}} \\
\midrule
TangentConv~\cite{tatarchenko2018tangent}       & 40.9 & -    & -    & -    & -    & 46.8 \\
TextureNet~\cite{huang2019texturenet}        & 54.8 & -    & -    & -    & -    & 63.0 \\
ScanComplete~\cite{dai2018scancomplete}      & 56.6 & -    & -    & -    & -    & 44.9 \\
Mix3D~\cite{nekrasov2021mix3d}             & 73.6 & -    & -    & -    & -    & -    \\
VMNet~\cite{hu2021vmnet}             & 73.2 & -    & -    & -    & -    & 67.2 \\
PTv3~\cite{wu2024point}     & 77.5    & -    & 80.4 & -    & -    & -    \\
MinkowskiNet~\cite{choy20194d}      & 69.0 & 77.5 & 78.0 & 83.7 & 54.2 & 64.6 \\
\midrule
\multicolumn{7}{c}{\textit{Zero-shot}} \\
\midrule
MSeg Voting~\cite{lambert2020mseg}      & 45.6 & 31.0 & 31.0 & 36.9 & 33.4 & 39.0 \\
CLIP2Scene~\cite{chen2023clip2scene}             & 25.1 & -    & -    & -    & -    & -    \\
CLIP-FO3D~\cite{zhang2023clip}             & 30.2 & 49.1 &  -    &  -    & -     &  -    \\
OpenScene(LS)$\ddag$~\cite{peng2023openscene}    & 54.2 & 66.6 & 36.7 & 42.7 & 43.4 & 53.5 \\
OpenScene(OS)$\ddag$~\cite{peng2023openscene} & 47.5 & 70.7 & 42.1 & 61.8 & 42.6 & 59.2 \\
GGSD~\cite{wang2024open}                  & 56.5 & 68.6 & 46.1 & 59.2 & -    & -    \\
PLA~\cite{ding2023pla}                    & 17.7 & 33.5 & -    & -    &   -   &   -   \\
GeoZe~\cite{mei2024geometrically}  & 55.8 & - & - & - & - & - \\
OV3D~\cite{jiang2024open}  & 57.3 & 72.9 & 44.6 & -    & 45.8 & 62.4 \\
PGOV3D~\cite{zhang2025pgov3d} & 59.5 & 73.2 & - & - & - & - \\
Diff2Scene~\cite{zhu2024open}             & 48.6 & -    & -    & -    & 45.5 & -    \\
DMA(OS)~\cite{li2024dense}           & 53.3 & 70.3 & 45.1 & - & 45.1 & 57.6 \\
CUA-O3D$\ddag$~\cite{li2025cross}           & 55.3 & 65.6 & -    & -    & 42.2 & 50.9 \\
SAS(stage1)~\cite{li2025sas}        & 59.2 & -    & 45.4 & -    & 46.3 & -    \\
SAS(stage2)~\cite{li2025sas}            & 61.9 & -    & 47.5 & -    & 48.6 & -    \\
\textbf{Ours}(OS)            & 53.4 & 74.8 & 47.5 & 67.3 & 47.7 & 66.1 \\  
\textbf{Ours}(LS)          & 59.8 & 72.5 & 40.4 & 48.3 & 47.8 & 60.5 \\
\textbf{Ours}($\text{SAS}^*$)              & \textbf{64.8} & \textbf{77.3} & \textbf{50.3} & \textbf{74.2} & \textbf{51.9} & \textbf{66.3} \\
\bottomrule
\end{tabular}
\vspace{-1.2em}
\end{table}

\begin{table*}
\small
\renewcommand{\arraystretch}{0.95}
\centering
\caption{\textbf{Long-tail scenario evaluation on Matterport3D} with varying numbers of categories. $\dag$ denotes results reproduced by us using officially provided weights, $\ddag$ denotes the use of the 2D-3D ensemble strategy. $\text{SAS}^*$ denotes using the same 2D features as in SAS~\cite{li2025sas}.}
\label{tab:matterport_compare}
\vspace{-1.0em}
\begin{tabular}{l *{8}{c}}
\toprule
\multirow{2}{*}{{Method}} & \multicolumn{2}{c}{{K = 21}} & \multicolumn{2}{c}{{K = 40}} & \multicolumn{2}{c}{{K = 80}} & \multicolumn{2}{c}{{K = 160}} \\
\cmidrule(lr){2-3} \cmidrule(lr){4-5} \cmidrule(lr){6-7} \cmidrule(lr){8-9}
& mIoU & mAcc & mIoU & mAcc & mIoU & mAcc & mIoU & mAcc \\
\midrule
OpenScene(LS)$\ddag$\dag~\cite{peng2023openscene}  & 43.4 & 53.5 & 25.6 & 31.3 & 12.5 & 15.8 & 6.4  & 8.2  \\
OpenScene(LS)\dag~\cite{peng2023openscene}  & 41.9 & 51.1 & 25.4 & 30.7 & 12.0 & 15.1 & 5.9  & 7.6  \\
OpenScene(OS)$\ddag$\dag~\cite{peng2023openscene}  & 42.6 & 58.7 & 34.4 & 49.9 & 21.0 & 33.2 & 10.9  & 18.3  \\
OpenScene(OS)\dag~\cite{peng2023openscene}  & 41.1 & 55.2 & 33.4 & 46.7 & 18.1 & 27.2 & 8.9  & 14.0  \\
DMA(OS)~\cite{li2024dense}          & 45.1 & 57.6 & 37.9 & 47.7 & 19.7 & 26.7 & 9.4  & 14.1 \\
SAS\dag~\cite{li2025sas}                     & 48.6 & 59.0 & 22.5 & 31.2 & 12.3 & 17.2 & 6.1  & 8.6  \\
\textbf{Ours}(LS)                               & 47.8 & 60.5 & 33.7 & 41.5 & 18.4 & 23.6 & 9.6  & 12.4 \\
\textbf{Ours}(OS)                            & 47.7 & 66.1 & \textbf{38.5} & \textbf{52.6} & \textbf{22.0} & \textbf{34.8} & \textbf{11.6} & \textbf{19.7} \\
\textbf{Ours}($\text{SAS}^*$)                                & \textbf{51.9} & \textbf{66.3} & 33.8 & 42.3 & 19.9 & 24.5 & 9.3  & 12.2 \\
\bottomrule
\end{tabular}
\vspace{-1.0em}
\end{table*}

\subsection{Main Results}
\textbf{Open-Vocabulary 3D Semantic Segmentation.}
We evaluate \textbf{GeoGuide} under the annotation-free open-vocabulary 3D semantic segmentation setting and compare it with representative fully-supervised and zero-shot methods on ScanNet v2~\cite{dai2017scannet}, Matterport3D~\cite{chang2017matterport3d}, and nuScenes~\cite{caesar2020nuscenes}.
As shown in Table~\ref{tab:multi_dataset_compare}, GeoGuide consistently outperforms prior zero-shot approaches across diverse datasets and 2D feature settings. When equipped with strong $\text{SAS}^*$ features, our method achieves state-of-the-art performance of 64.8 mIoU on ScanNet v2, 50.3 mIoU on nuScenes, and 51.9 mIoU on Matterport3D. These results represent substantial improvements of +5.6, +4.9, and +5.6 mIoU over the powerful SAS (stage1) baseline, respectively. Remarkably, our approach even narrows the gap with fully-supervised methods, achieving results comparable to or exceeding early fully-supervised approaches like TangentConv and TextureNet on ScanNet v2 while requiring no 3D annotations. This significant leap is attributed to our framework's ability to correct geometric inconsistencies inherited from 2D distillation through hierarchical consistency modeling. A key strength of GeoGuide lies in its remarkable generalization across different 2D feature extractors. With LSeg features, GeoGuide surpasses OpenScene (LSeg)-2D3D by +5.6 mIoU on ScanNet v2. With OpenSeg, the gains are equally impressive: +5.9 mIoU on ScanNet v2 and +5.4 mIoU on nuScenes over OpenScene (OpenSeg)-2D3D. This consistent performance elevation across diverse 2D features highlights that \textbf{our framework addresses fundamental geometric inconsistency issues rather than merely exploiting specific 2D feature characteristics.} Beyond mIoU, GeoGuide also achieves significant gains in mAcc, particularly on nuScenes (+5.5 mAcc over OpenScene with OpenSeg) and ScanNet v2 (+5.9 mAcc over OpenScene with LSeg). The dual improvement in both mIoU and mAcc indicates that our method not only increases segmentation coverage but also enhances per-class prediction accuracy. This suggests that geometric consistency modeling helps the network learn more discriminative and class-specific features, rather than simply improving boundary localization. 
In summary, these comprehensive results validate that leveraging pretrained 3D geometric priors to enforce hierarchical consistency from local to global is both essential and effective for addressing the fundamental limitations of conventional 2D-to-3D knowledge transfer in open-vocabulary 3D semantic segmentation.

\noindent\textbf{Long-Tail Scenario Evaluation.}
To  further evaluate the robustness of GeoGuide in handling a large number of categories, we  conduct a long-tail scenario evaluation on the Matterport3D test set using top-K categories ( K = 21, 40, 80, 160). Under the zero-shot setting, the same model trained on the Matterport3D training set is directly evaluated across all K values.
As shown in Table~\ref{tab:matterport_compare},  GeoGuide consistently achieves state-of-the-art performance under all category settings. Although all methods experience performance drops as K increases, which is expected in long-tail scenarios, GeoGuide exhibits a much smaller performance drop, reflecting its superior discriminative capability in fine-grained category recognition. 
This robustness stems from our proposed inter-instance relation consistency module, which reinforces semantic coherence among the same-category instances and maintains reliable discrimination even as inter-class similarity increases.

\begin{table*}
\centering
\small
\renewcommand{\arraystretch}{0.95}
\caption{\textbf{Cross-domain generalization evaluation.} Models are \textbf{trained on ScanNet v2 and evaluated on Matterport3D (21/40/80/160 categories)}, $\dag$ denotes results reproduced by us using officially provided weights, $\ddag$ denotes the use of the 2D-3D ensemble strategy. $\text{SAS}^*$ denotes using the same 2D features as in SAS~\cite{li2025sas}.}
\vspace{-1.0em}
\label{tab:Cross_domain_generalization}
\begin{tabular}{l *{8}{c}}
\toprule
\multirow{2}{*}{{Method}} & \multicolumn{2}{c}{{K = 21}} & \multicolumn{2}{c}{{K = 40}} & \multicolumn{2}{c}{{K = 80}} & \multicolumn{2}{c}{{K = 160}} \\
\cmidrule(lr){2-3} \cmidrule(lr){4-5} \cmidrule(lr){6-7} \cmidrule(lr){8-9}
& mIoU & mAcc & mIoU & mAcc & mIoU & mAcc & mIoU & mAcc \\
\midrule
OpenScene(OS)\dag~\cite{peng2023openscene} & 34.4   & 50.8   & 23.8   & 37.6   & 12.9   & 21.0   & 6.3    & 11.1   \\
OpenScene(LS)\dag~\cite{peng2023openscene}  & 36.4   & 48.1   & 21.3   & 27.7   & 10.8   & 14.1   & 5.4    & 7.1   \\
GGSD~\cite{wang2024open}                          & 40.1   & 54.4   & 22.8    & 31.6    & 11.9    & 16.2    & 6.3     & 9.6     \\
CUA-O3D$\ddag$~\cite{li2025cross}                  & 37.4   & 49.2   & 23.3    & 30.2    & 12.2    & 16.3    & 6.1     & 8.4     \\
SAS\dag~\cite{li2025sas}                           & 43.0  & 56.5  & 18.0   & 26.2   & 9.8    & 14.7   & 4.9   & 7.3   \\
\textbf{Ours}(LS)                     & 47.2  & 61.6  & 31.3   & 39.3   & 16.9 & 22.6 & 8.4  & 11.3 \\
\textbf{Ours}(OS)                     & 43.7  & 64.4  & \textbf{31.9}   & \textbf{46.6}   & \textbf{18.8} & \textbf{29.7} & \textbf{9.7}  & \textbf{17.5} \\
\textbf{Ours}($\text{SAS}^*$)                     & \textbf{50.1}  & \textbf{65.1}  & 31.7 & 39.8 & 17.7 & 23.9 & 8.8  & 12.1 \\

\bottomrule
\end{tabular}
\vspace{-1.0em}
\end{table*}

\begin{table}
\centering
\small
\renewcommand{\arraystretch}{0.95}
\caption{\textbf{Cross-domain generalization evaluation}. Models are \textbf{trained on Matterport3D and evaluated on ScanNet v2}. $\dag$ denotes results reproduced by us using officially provided weights, $\ddag$ denotes the use of the 2D-3D ensemble strategy. $\text{SAS}^*$ denotes using the same 2D features as in SAS~\cite{li2025sas}.}
\label{tab:Cross_domain_generalization_2}
\vspace{-1em}
\begin{tabular}{lcc}
\toprule
\textbf{Method}                & mIoU  & mAcc  \\
\midrule
OpenScene(LS)\dag~\cite{peng2023openscene}    & 37.2 & 44.8 \\
OpenScene(OS)\dag~\cite{peng2023openscene} & 38.8 & 52.2 \\
CUA-O3D$\ddag$~\cite{li2025cross}        & 38.6  & 46.6  \\
SAS\dag~\cite{li2025sas}                   & 44.3 & 52.9 \\
\textbf{Ours}(LS)            & 52.2 & 64.5 \\
\textbf{Ours}(OS)         & 51.1 & \textbf{71.3} \\
\textbf{Ours}($\text{SAS}^*$)       & \textbf{57.2} & 69.9 \\
\bottomrule
\end{tabular}
\vspace{-0.2em}
\end{table}

\noindent\textbf{Cross-Domain Generalization.}
To evaluate cross-domain robustness, we conduct bidirectional transfer experiments without fine-tuning: (1) training on ScanNet v2 and evaluating on Matterport3D across varying scales (K = 21, 40, 80, 160) in Table~\ref{tab:Cross_domain_generalization}, and (2) the reverse direction  from Matterport3D to ScanNet v2 in Table~\ref{tab:Cross_domain_generalization_2}.
Our method consistently outperforms all baselines across both directions and all category scales. Notably, it exhibits superior robustness as task complexity increases, maintaining stable performance even in challenging high-category scenarios. This consistent superiority in both transfer directions demonstrates that our hierarchical geometry-guided framework effectively captures domain-invariant geometric structures through universal geometric priors that provide stable spatial cues independent of appearance variations across datasets.

\begin{figure}
    \centering
    \includegraphics[width=1.0\linewidth, keepaspectratio]
    {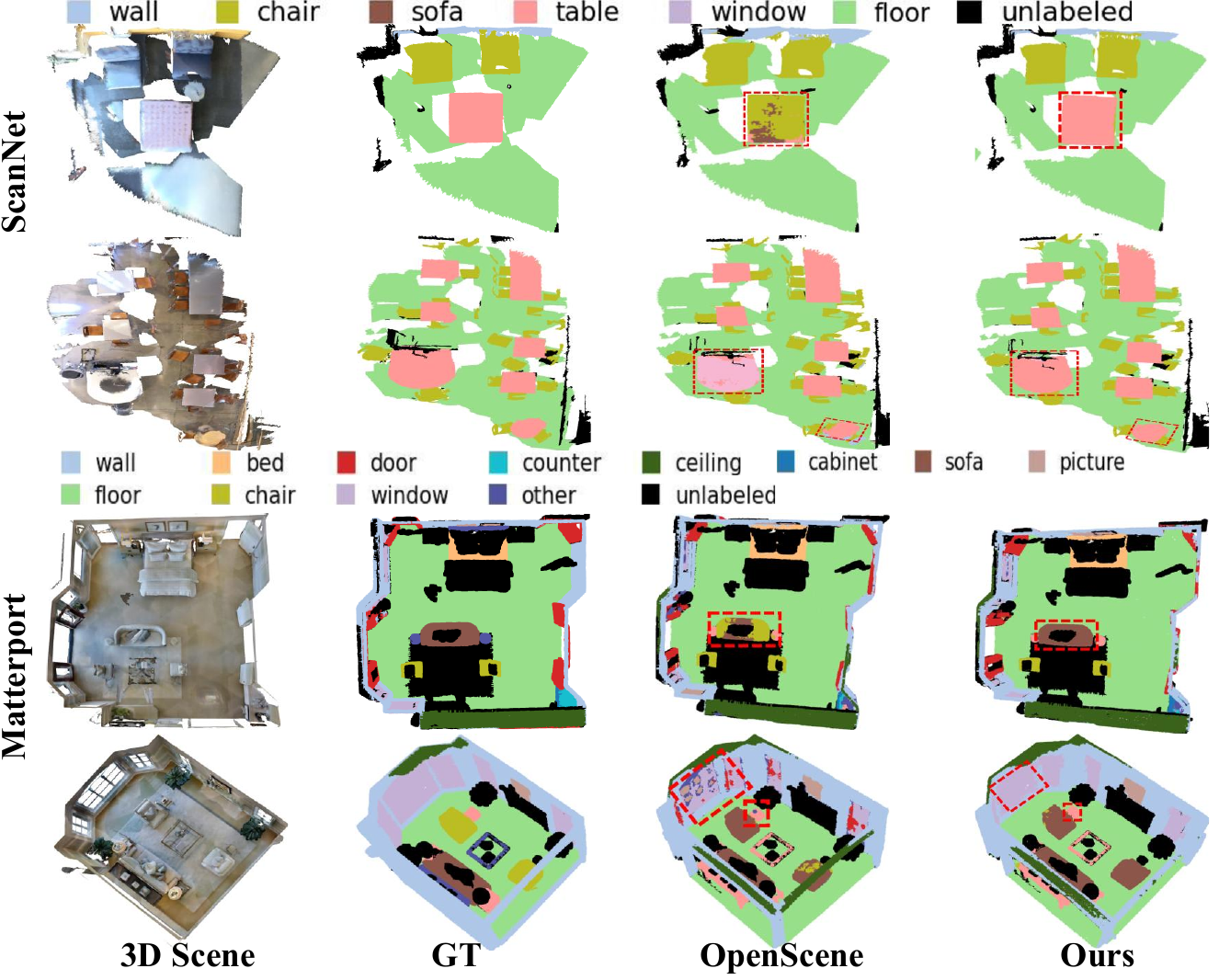}
     \vspace{-2em}
     \caption{\textbf{Visualization of semantic segmentation results } on ScanNet v2 and Matterport3D.} 
       \vspace{-1.2em}
    \label{fig:qualitative}
    
\end{figure}

\noindent\textbf{Qualitative Comparison.}
To visually assess the segmentation quality, we present qualitative results of our approach and the baseline on ScanNet v2 and Matterport3D in Figure~\ref{fig:qualitative}. \textbf{GeoGuide} better preserves instance boundaries, completeness, and fine-grained semantic consistency. It effectively corrects boundary ambiguities and occlusion-induced errors that affect baseline methods, resulting in more accurate and perceptually coherent 3D segmentations.

\begin{table}
\centering
\small
\renewcommand{\arraystretch}{0.95}
\caption{\textbf{Ablation study of the model with different designs}.}
\label{tab:geoguide_ablation}
\vspace{-1.0em}
\begin{tabular}{c cccc cc}
\toprule
{Group} & {SD} & {USD} & {IMR} & {IIRC} & {mIoU} & {mAcc} \\
\midrule
\textbf{A} & \xmark & \xmark & \xmark & \xmark & 54.2 & 66.6 \\
\textbf{B} & \xmark & \xmark & \xmark & \xmark & 53.1 & 63.9 \\
\textbf{C} & \xmark & \xmark & \xmark & \xmark & 53.9 & 65.8 \\
\midrule
\textbf{D} & \cmark & \xmark & \xmark & \xmark & 55.8 & 67.2 \\
\textbf{E} & \xmark & \cmark & \xmark & \xmark & 56.4 & 68.3 \\
\textbf{F} & \xmark & \xmark & \cmark & \xmark & 56.8 & 68.9 \\
\textbf{G} & \xmark & \cmark & \cmark & \xmark & 57.9 & 70.5 \\
\textbf{H} & \xmark & \xmark & \xmark & \cmark & 57.1 & 69.3 \\
\midrule
\textbf{I} & \xmark & \cmark & \cmark & \cmark & \textbf{59.8} & \textbf{72.5} \\
\bottomrule
\end{tabular}%
\vspace{-1.0em}
\end{table}

\subsection{Ablation Studies and Analysis}
In this section, we conduct comprehensive ablation studies under the zero-shot setting on the ScanNet v2 validation set to validate  the effectiveness of each component of our framework. All ablation experiments employ LSeg~\cite{li2022language} as the 2D feature extractor.

\noindent\textbf{Impact of Pretrained Model.}
As shown in Table~\ref{tab:geoguide_ablation}, [A] denotes  the performance of our baseline~\cite{peng2023openscene}. To analyze the effect of pretraining, we conduct experiments [B] and [C]. Specifically, [B] retrains the pretrained backbone~\cite{wu2025sonata} from scratch without loading pretrained weights, while [C] directly replaces the original OpenScene backbone with the pretrained one~\cite{wu2025sonata}. The comparison between [B] and [C] shows that pretraining alone yields only marginal improvements. We conjecture that without explicit  geometric constraints, the geometric priors learned during pretraining gradually deteriorate and fail to fully exploit the potential of the pretrained model. This phenomenon confirms the main issue highlighted in the introduction, namely that directly introducing 3D pretrained features can instead lead to degradation of geometric information.

\noindent\textbf{Module Effectiveness Analysis.}
To assess the contribution of each proposed component, we conduct experiments by adding different modules individually  or their combinations during 3D model training, as summarized  in Table~\ref{tab:geoguide_ablation}. Here, SD denotes the superpoint distillation method (average pooling) as adopted in prior work such as SAS~\cite{li2025sas} and GGSD~\cite{wang2024open}. Comparing [D] and [E], our uncertainty-weighted distillation effectively suppresses noisy features and emphasizes reliable superpoint representations, outperforming simple average pooling. Moreover, [E], [F], and [H] show that incorporating any geometry-semantic consistency constraint module helps mitigate the degradation of pretrained geometric priors and improves segmentation performance. Finally, [I] demonstrates that combining all modules achieves the best performance, further surpassing all individual module configurations. 
These results indicate that the three modules work synergistically, forming a hierarchical geometry-semantic consistency constraint from local to global levels, and fully unlock the potential of pretrained geometric priors. For more ablation studies, please refer to supplementary material.


\section{Conclusion}
In this paper, we present GeoGuide, a hierarchical geometry-guided framework for open-vocabulary 3D semantic segmentation. Our approach addresses the challenge of preserving intrinsic 3D geometric priors during 2D-to-3D knowledge distillation, which has been overlooked in prior work.
Specifically, the proposed uncertainty-based superpoint distillation module mitigates noisy 2D supervision and enhances intra-superpoint consistency. The instance-level mask reconstruction module leverages geometric priors to recover complete instance structures. Finally, the inter-instance relation consistency module aligns geometric and semantic similarities across instances to handle viewpoint variations.
Extensive experiments on multiple benchmarks validate the effectiveness of GeoGuide.

\section{Acknowledgements}
This work was partially supported by the National Natural Science Foundation of China (42595545) and the Youth Innovation Promotion Association of CAS.

{
    \small
    \bibliographystyle{ieeenat_fullname}
    \bibliography{main}
}


\end{document}